\documentclass[letterpaper, 10 pt, conference]{ieeeconf}

\IEEEoverridecommandlockouts                              

\usepackage[T1]{fontenc}

\usepackage{tikz} 

\usepackage{amsfonts}	
\usepackage{amsmath}	
\usepackage{amssymb}    
\usepackage{siunitx}
\usepackage{pifont}   

\usepackage{booktabs}
\usepackage{makecell}  
\usepackage[flushleft]{threeparttable}  
\usepackage{multirow}
\usepackage{rotating}

\usepackage{xspace}    
\usepackage{xcolor}    
\let\labelindent\relax
\usepackage[inline]{enumitem} 
\usepackage{textcomp}  
\usepackage{gensymb}   
\usepackage{graphicx} 
\usepackage[caption=false,font=footnotesize]{subfig}
\usepackage{microtype}
\usepackage{cite}
\usepackage{cuted}

\makeatletter
\let\NAT@parse\undefined
\makeatother

\usepackage{url}
\usepackage[pdfencoding=auto, hidelinks]{hyperref} 
\usepackage[hang,flushmargin]{footmisc}

\usepackage[capitalize]{cleveref}
\crefname{section}{Sec.}{Secs.}
\Crefname{section}{Section}{Sections}
\Crefname{table}{Table}{Tables}
\crefname{table}{Tab.}{Tabs.}

\usepackage[nogroupskip,acronyms,nopostdot,style=super,nonumberlist,toc]{glossaries}

\usepackage{colortbl}
\newcommand{\greyrule}{\arrayrulecolor{black!30}\midrule\arrayrulecolor{black}}

\newcommand{\cmark}{\ding{51}}  

\newacronym{ad}{AD}{automated driving}
\newacronym{hd}{HD}{high-definition}
\newacronym{fv}{FV}{frontal view}
\newacronym{bev}{BEV}{bird's-eye-view}
\newacronym{slam}{SLAM}{simultaneous localization and mapping}
\newacronym{vqa}{VQA}{visual question answering}
\newacronym{vi}{VI}{visual-inertial}
\newacronym{pgo}{PGO}{pose graph optimization}

\begin{document}

\title{\LARGE \bf
Collaborative Dynamic 3D Scene Graphs for Automated Driving
}

\author{
Elias Greve$^{1*}$,
Martin Büchner$^{1*}$,
Niclas Vödisch$^{1*}$,
Wolfram Burgard$^{2}$,
and Abhinav Valada$^{1}$
\thanks{$^{*}$ Equal contribution.}%
\thanks{$^{1}$ Department of Computer Science, University of Freiburg, Germany.}%
\thanks{$^{2}$ Department of Eng., University of Technology Nuremberg, Germany.}%
\thanks{This work was funded by the European Union’s Horizon 2020 research innovation program grant No 871449-OpenDR, the German Research Foundation (DFG) Emmy Noether Program grant No 468878300, and an academic grant from NVIDIA.}%
\thanks{Accepted for the 2024 IEEE Int. Conf. on Robotics and Automation.}
}


\maketitle


\begin{abstract}
    Maps have played an indispensable role in enabling safe and automated driving. Although there have been many advances on different fronts ranging from SLAM to semantics, building an actionable hierarchical semantic representation of urban dynamic scenes and processing information from multiple agents are still challenging problems. 
In this work, we present Collaborative URBan Scene Graphs \mbox{(CURB-SG)} that enable higher-order reasoning and efficient querying for many functions of automated driving. CURB-SG leverages panoptic LiDAR data from multiple agents to build large-scale maps using an effective graph-based collaborative SLAM approach that detects inter-agent loop closures.
To semantically decompose the obtained 3D map, we build a lane graph from the paths of ego agents and their panoptic observations of other vehicles. Based on the connectivity of the lane graph, we segregate the environment into intersecting and non-intersecting road areas. Subsequently, we construct a multi-layered scene graph that includes lane information, the position of static landmarks and their assignment to certain map sections, other vehicles observed by the ego agents, and the pose graph from SLAM including 3D panoptic point clouds.
We extensively evaluate CURB-SG in urban scenarios using a photorealistic simulator. We release our code at \mbox{\url{http://curb.cs.uni-freiburg.de}}.

\end{abstract}
\glsresetall




\section{Introduction}
Spatial and semantic understanding of the environment is crucial for the safe and autonomous navigation of mobile robots and self-driving cars. Recent autonomy systems leverage \gls{hd} map information as effective priors for several downstream tasks in \gls{ad} including perception~\cite{yang2018hdnet}, localization~\cite{cattaneo2024cmrnext}, planning~\cite{diazdiaz2022hd}, and control~\cite{trumpp2023efficient}.
\gls{hd} maps are often constructed and maintained in a top-down manner~\cite{poggenhans2018lanelet2}, i.e., relying on traffic authorities or via arduous labeling efforts. In contrast, automatic bottom-up \gls{ad} mapping approaches show high accuracy \cite{chen2019sumapp, koide2019a} while being limited to occupancy or semantic mapping using, e.g., dense voxel grid manifolds.
With respect to \gls{ad}, map representations should ideally fulfill the following requirements~\cite{wong2020mapping}:
\begin{enumerate*}[topsep=0pt]
    \item completeness and accuracy while scaling to large areas;
    \item frequent updates to capture structural changes;
    \item higher-level topological information grounded in rich sensor data;
    \item efficient access and information querying.
\end{enumerate*}
Given these requirements, typical SLAM maps only enable classical spatial or point-level semantic querying. We envision that modern \gls{ad} mapping approaches should provide the means to process vision and language queries, e.g., from foundation models~\cite{clip}. Enabling such demands can only become feasible by abstracting from given maps using sparse representations.

\begin{figure}[t]
    \centering
    \includegraphics[width=\linewidth]{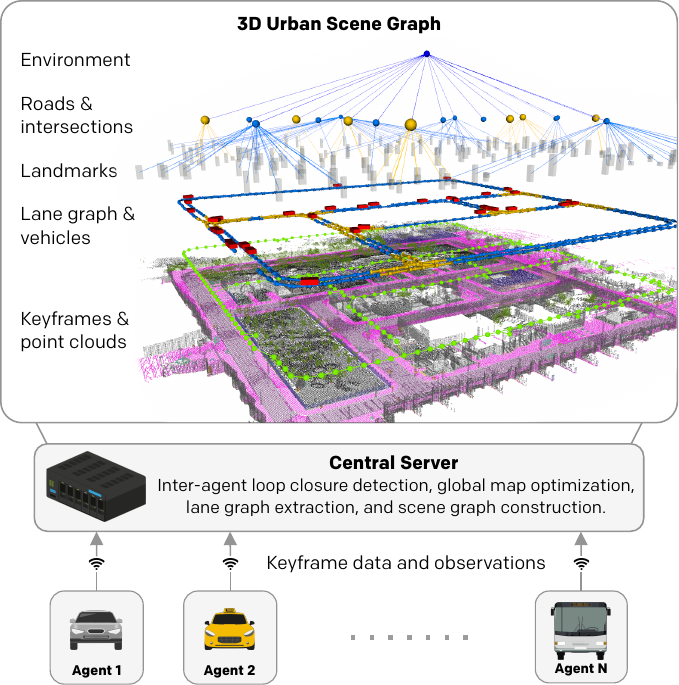}
    \caption{For our proposed collaborative urban scene graphs (CURB-SG), multiple agents send keyframe packages with their local odometry estimates and panoptic LiDAR scans to a central server that performs global graph optimization. We subsequently partition the environment based on a lane graph from agent paths and other detected cars. Together with the 3D map, the lane graph forms the base of the large-scale hierarchical scene graph.}
    \label{fig:cover}
    \vspace{-.5cm}
\end{figure}

In this work, we propose \textit{Collaborative URBan Scene Graphs} (CURB-SG) that effectively address the aforementioned requirements by constructing a hierarchical graph structure of the environment as shown in \cref{fig:cover}. 3D scene graphs enable efficient data storage of large environments while being queryable and preserving spatial information. Previous works on 3D scene graphs~\cite{bavle2022_plus, hughes2022hydra, armeni20193d} focus on indoor environments, whose taxonomy cannot be directly transferred to large-scale urban domains. To close this gap, we introduce the following analogy to indoor variants: Cities (buildings) can be separated into intersections and roads (rooms), which contain static landmarks such as traffic signs (furniture) as well as dynamic objects such as vehicles (humans). We enable this partitioning by generating an online lane graph that serves as a common link among multiple graph layers.
Addressing frequent updates and multi-agent cooperation, our method leverages a centralized collaborative SLAM approach that combines panoptic LiDAR data and local odometry estimates into a single 3D map while optimizing a global pose graph that benefits from inter-agent loop closures. Following the spirit of previous works on scene graphs~\cite{bavle2022_plus, hughes2022hydra, armeni20193d}, we extensively evaluate our proposed method on simulated data using the CARLA simulator~\cite{dosovitskiy2017carla}.

To summarize, the main contributions are as follows:
\begin{enumerate}[topsep=0pt]
    \item We introduce a novel algorithm for representing urban driving environments as dynamic 3D scene graphs that are constructed from multi-agent observations to efficiently cover large areas.
    \item We demonstrate an effective partitioning of urban environments using lane graphs constructed on the fly from panoptic \mbox{LiDAR} observations in a cooperative manner.
    \item We present an efficient collaborative graph SLAM method to continuously update semantic maps while addressing scalability via edge contraction.
    \item We provide extensive evaluations of the building blocks of our proposed framework.
    \item We make our code and sample data publicly available at \mbox{\url{http://curb.cs.uni-freiburg.de}}.
\end{enumerate}

\section{Related Work}
\label{sec:related-work}

In this section, we first present a summary of LiDAR-based odometry and mapping, followed by an overview of multi-agent SLAM, and scene graphs in \acrfull{ad}.\looseness=-1


{\parskip=3pt
\noindent\textit{LiDAR SLAM:}
LiDAR-based mapping has been pioneered by LOAM~\cite{zhang2014loam} that estimates robot motion from scan registration via ICP between subsequent point clouds.
To address the full SLAM problem, HDL Graph SLAM~\cite{koide2019a} combines LiDAR odometry with local loop closure detection and performs joint pose graph optimization.
Leveraging semantic segmentation, SUMA++~\cite{chen2019sumapp} masks dynamic classes during the mapping stage and proposes a semantic-aided variant of ICP.
\mbox{PADLoC}~\cite{arce2023padloc} exploits panoptic segmentation during training to stabilize both loop closure detection and registration.
In this work, we use panoptic point clouds to create a large-scale semantic 3D map forming the base layer of our scene graph.
}


{\parskip=3pt
\noindent\textit{Collaborative SLAM:}
To cover large environments and to increase mapping speed, SLAM research begins to shift towards multi-agent methods~\cite{zou2019collaborative}. Generally, collaborative SLAM can be realized in a centralized or distributed manner.
Initial works such as C\textsuperscript{2}TAM~\cite{riazuelo2014c2tam} belong to the centralized category, performing global bundle adjustment on a server and localization on the clients. A similar paradigm is adopted by CVI-SLAM~\cite{Karrer2018cvislam} and COVINS~\cite{schmuck2021covins}, proposing \gls{vi} SLAM systems for a fleet of UAVs. While the robots run local \gls{vi} odometry, a central server collects this information, searches for inter-agent loop closures to perform global optimization, and removes redundant data. With respect to LiDAR SLAM, LAMP~2.0~\cite{chang2022lamp2} allows collaboration between different types of robots to map large-scale underground environments.
A similar use case is addressed by Swarm-SLAM~\cite{lajoie2023swarmslam}, which supports further sensor modalities. Following a distributed paradigm, information is directly shared between the agents using peer-to-peer communication. Kimera-Multi~\cite{tian2022kimeramulti} is a \gls{vi} SLAM method that includes semantic information in the generated 3D mesh. For data fusion, it employs distributed \gls{pgo}. Finally, DisCo-SLAM~\cite{huang2022discoslam}  proposes a LiDAR-based approach addressing the initially unknown relative position of the agents. For this, they use Scan Context~\cite{kim2018scan} descriptors for global loop closure detection without spatial priors.
In this work, we follow the centralized paradigm since we leverage collaborative SLAM to generate a single consistent scene graph that can be made available to other traffic participants to query information.
}


{\parskip=3pt
\noindent\textit{Scene Graphs for Automated Driving:}
3D scene graphs constitute an effective interface unifying pose graphs from large-scale mapping and local information~\cite{sgraphs_2022} such as frame-wise object detections~\cite{lang2022robust}, topological mapping~\cite{buchner2023learning, he_wacv}, or semantic segmentation~\cite{gosala2023skyeye, mohan2022amodal}. Additionally, graphs enable the structural disassembly of large-scale scenes into objects and their relationships and facilitate higher-level reasoning, e.g., in the vision and language domain~\cite{visual_genome}. This further allows for efficient hierarchical abstraction in both spatial and semantic regimes~\cite{armeni20193d, rosinol20203DDS}. 
So far, 3D scene graphs for environment representation have only been applied in indoor domains. The first work in this field~\cite{armeni20193d} proposes an offline, multi-layered hierarchical representation based on RGB images.
Kim~\textit{et~al.}~\cite{kim_cyber} were the first to generate 3D scene graphs from RGB-D images for visual question answering~\cite{vqa} and task planning. Using a learning-based pipeline, Wald~\textit{et~al.}~\cite{wald_3dssg} construct a 3D scene graph from an instance-segmented point cloud while predicting node and edge semantics in an offline manner. Rosinol~\textit{et~al.}~\cite{rosinol20203DDS} present an offline framework capable of generating hierarchical scene graphs from dynamic indoor scenes that are divided into buildings, rooms, places, objects, and agents, as well as a metric-semantic mesh. Different from the aforementioned frameworks, Hydra~\cite{hughes2022hydra}, SceneGraphFusion~\cite{wu_scenegraphfusion}, and \mbox{S-Graphs}~\cite{sgraphs_2022} present real time-capable approaches.
While Hydra does not tightly couple the optimized pose graph with the 3D scene graph, the non-hierarchical \mbox{S-Graphs}~\cite{sgraphs_2022} close this gap. The follow-up work \mbox{S-Graphs+}~\cite{bavle2022_plus} also encodes hierarchies.
In this work, we combine collaborative SLAM and 3D scene graphs to build hierarchical maps for \gls{ad}. To the best of our knowledge, our work constitutes the first approach to 3D scene graph construction of urban driving scenes with a tightly coupled integration of inter-agent loop closures. Furthermore, we show how multi-agent cooperation facilitates frequent map updates and completeness.
}

\section{Technical Approach}

\begin{figure*}[t]
    \centering
    \includegraphics[width=1.0\textwidth]{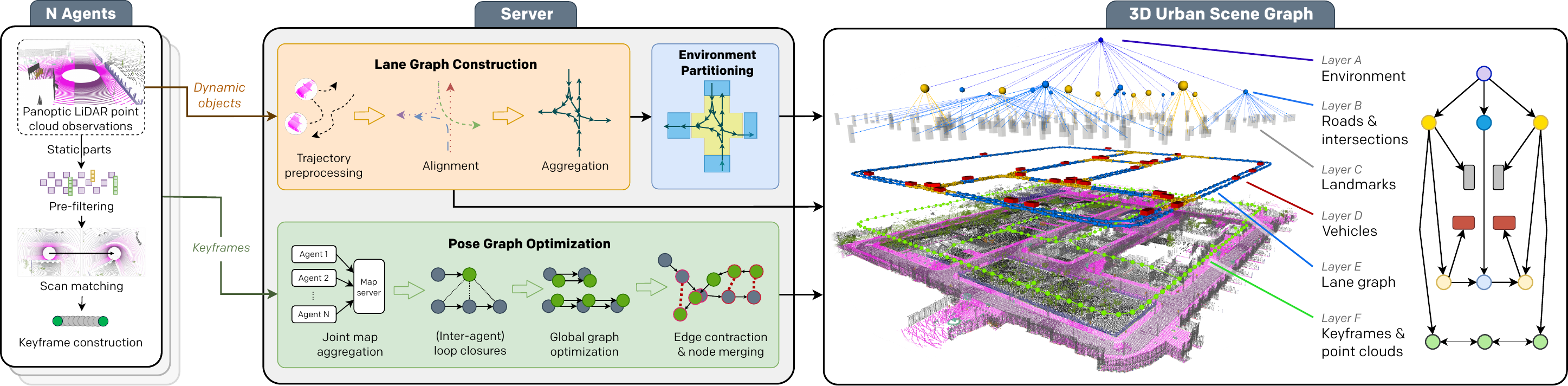}
    \vspace*{-.5cm}
    \caption{Overview of CURB-SG: Multiple agents obtain panoptically segmented LiDAR data and provide an odometry estimate based on the static parts of the point cloud. A centralized server instance then performs \acrfull{pgo} including inter-agent loop closure detection and edge contraction based on the agents' inputs. Tightly coupled to the pose graph, we aggregate a lane graph from panoptic observations of other vehicles as well as the agent's trajectories.
    Next, the lane graph is partitioned to retrieve a topological separation that allows for the hierarchical abstraction of larger environments.
    }
    \label{fig:overview}
    \vspace*{-.4cm}
\end{figure*}

In this section, we present our \mbox{CURB-SG} approach for collaborative urban scene graphs. As illustrated in \cref{fig:overview}, CURB-SG is comprised of several components. In \mbox{\cref{ssec:collaborative-slam}}, we describe our approach for collaborative SLAM to effectively combine panoptic information. Here, multiple agents transmit their onboard LiDAR odometry estimates along with panoptic point clouds to a central compute unit. This server combines the data by detecting intra- and inter-agent loop closures and performs \acrfull{pgo} to generate a globally consistent 3D map. In \cref{ssec:scene-graph}, we propose to further aggregate the paths of the agents and other observed vehicles to extract an online lane graph allowing for partitioning the city into intersections and roads. Finally, the server registers dynamic traffic participants on the lane graph and generates a hierarchical scene graph by assigning static landmarks to the closest intersection or road.


\subsection{Collaborative SLAM}
\label{ssec:collaborative-slam}

We leverage collaborative LiDAR SLAM as the backend in our proposed CURB-SG. Due to its reliable performance and well-maintained code base, we build on top of HDL Graph SLAM~\cite{koide2019a} and extend it to a multi-agent scenario following a centralized approach as described in \cref{sec:related-work}.
In this section, we describe the steps performed by each agent, followed by the centralized \gls{pgo} as depicted in \cref{fig:overview}. Finally, we provide further details on how CURB-SG explicitly addresses both long-term and large-scale mapping.


{\parskip=3pt
\noindent\textit{Agents:} 
Each agent is equipped with a LiDAR sensor to capture sparse 3D point clouds, which contain spatial information as well as point-wise panoptic segmentation labels. Initially, a point cloud is separated into its static and dynamic components following the conventional categorization of ``stuff'' and ``thing'' classes~\cite{sirohi2022efficientlps}. Similar to SUMA++~\cite{chen2019sumapp}, we extract the static points for creating the map.
In contrast to HDL Graph SLAM~\cite{koide2019a}, we utilize different voxel grid sizes for the various semantic classes. This approach retains more dense information where required, e.g., poles and traffic signs are being processed at a more fine-grained level than roads or buildings.
Next, we perform point cloud registration via \mbox{FAST-GICP}~\cite{koide2021voxelized} between subsequent LiDAR scans to estimate the motion of an agent. Following the common methodology and to reduce the required bandwidth between the agents and the server, we generate keyframes after a specified traveled distance based on LiDAR odometry. Each keyframe is sent to the server and contains an estimated pose and the static LiDAR point cloud with semantic labels, i.e., the ``stuff'' points. 
Since car instances (\textit{dynamic objects}) contribute to the online construction of a lane graph (see \cref{ssec:scene-graph}), the ``thing'' points from all the LiDAR scans are transformed relative to the pose of the previous keyframe and sent separately.
}


{\parskip=3pt
\noindent\textit{Server:}
The centralized server receives keyframes from all the agents and processes them as outlined in the green box of \cref{fig:overview}: 
First, upon receiving the first keyframe sent by an agent, the server registers this agent to the global pose graph.
Second, the server searches for loop closure candidates between the added keyframe and the existing nodes in the pose graph to find both intra- and inter-agent loop closures. We rely on the original loop closure detection technique of HDL Graph SLAM~\cite{koide2019a}, i.e., all nodes within a local search radius are considered to be candidates. If the fitness score of the ICP algorithm is below a threshold, a loop closure edge is added to the pose graph. Due to relying on an initial guess, we utilize the absolute ground truth value for the registration of a new agent. In practice, this could either be solved with GNSS measurements or by conducting an efficient global search for loop closure candidates leveraging point cloud descriptors~\cite{huang2022discoslam}.
Third, the server performs \gls{pgo} using g\textsuperscript{2}o~\cite{kuemmerle2011g2o} to integrate the newly added keyframes and detected loop closures.
To address scalability, we employ edge contraction as detailed in the following paragraph.
Finally, we apply the same semantics-based voxelization to the entire 3D map as performed by the agents on their local LiDAR scans.
}

\begin{figure}[t]
    \centering
    \includegraphics[width=.9\linewidth]{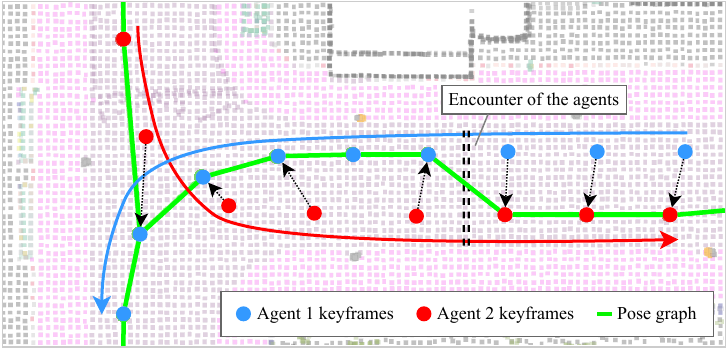}
    \vspace{-.3cm}
    \caption{In this example, two agents drive along the same road while passing each other at the dashed line. The detected loop closures yield additional edges in the pose graph. After optimization, the edges that carry redundant information are contracted by merging the older node into the more recently added node to update the map information.}
    \label{fig:edge-contraction}
   \vspace{-.3cm}
\end{figure}

{\parskip=3pt
\noindent\textit{Long-Term and Large-Scale Mapping:}
If not handled explicitly, the pose graph would continue to grow while the mapping progresses. Since every keyframe contains a 3D point cloud, this not only significantly slows down the \gls{pgo} but also increases memory consumption and disk storage. To address both problems, we remove the nodes and edges from the graph that carry redundant information \textit{(edge contraction \& node merging)}. In \cref{fig:edge-contraction}, two agents have driven along the same road yielding multiple loop closures. Using a heuristic-driven approach, the loop closure edges that carry redundant information are being contracted by merging nodes. By redirecting the edges of the omitted to the remaining node, we ensure the legal connectivity of the pose graph. Notably, this is done after the \gls{pgo} step. Consequently, the final pose graph becomes easier to maintain and more efficient to query when searching for new loop closures.
The point cloud data associated with a removed node is combined with the data of the persisting node while omitting older data to guarantee up-to-date map information. In contrast, the dynamic observations linked to a node are completely transferred as they contribute towards the construction of the lane graph explained in \cref{ssec:scene-graph}. For the same reason, each removed node is turned into a passive observation that stores the driven path of an ego agent.
}

\subsection{Scene Graph Generation}
\label{ssec:scene-graph}

The second key component of CURB-SG is a scalable environment representation of urban outdoor scenes for \gls{ad}. Besides the aforementioned 3D semantic map, CURB-SG constructs a tightly coupled hierarchical abstraction of the environment as shown in \cref{fig:overview}. By analogy with the separation of indoor scenes into buildings, rooms, and places~\cite{rosinol20203DDS, armeni20193d, hughes2022hydra}, we decompose a constructed lane graph into intersecting and non-intersecting road areas allowing for spatial and semantic abstraction.
 
The root of our CURB-SG representation is given by \textit{Layer~A} that holds environment/city-level information. This environment is then spatially divided into intersections and their connecting roads (\textit{Layer~B}), which serve as the categorical counterparts to rooms and corridors in indoor scenes. Since the partitioning of our environment is based on a lane graph (presented in \textit{Layer~E}), the connectivity of \textit{Layer~B} is implicitly given by the connectivity of the lane graph (colored segments, \cref{fig:overview}). Next, we map static landmarks such as traffic signs and poles contained in \textit{Layer~C} including their bounding box to their corresponding spatial area defined by \textit{Layer~B}. These landmarks can serve as priors for localization or object detection.
\textit{Layer~D} holds all currently observed dynamic vehicles. We map dynamic vehicles to their closest respective lane graph node, as defined in \textit{Layer~E}, to provide efficient access for downstream tasks, e.g., trajectory prediction.
Central to this approach, \textit{Layer~E} is a directed lane graph to encode the low-level topology for vehicle navigation and is inferred from the paths of the ego agents as well as other perceived vehicles. We provide further details in the next paragraph. The lane graph defines the connectivity of the different spatial regions in the urban environment, comparable to edges among rooms in indoor scene graph variants. Finally, \textit{Layer~F} contains the pose graph from our SLAM backend and encodes LiDAR data in the form of semantic point clouds. As discussed in \cref{ssec:collaborative-slam}, this layer is subject to continuous optimization and dynamic restructuring, e.g., due to loop closure detection and edge contraction. Based on the edges between the keyframes in this layer and spatial areas (\textit{Layer~B}), 3D map information is easily accessible given a rough road-level position estimate.


{\parskip=3pt
\noindent\textit{Lane Graph Construction:} 
We generate a lane graph of the environment leveraging the trajectories of the ego agents as well as observations of surrounding vehicles. As the LiDAR point clouds of the agents contain instance IDs, we are able to differentiate between multiple observed vehicle instances in the agents' surroundings. For each observed vehicle, we extract the centroid of its partial point cloud. The position of a centroid is stored relative to the most recent keyframe. After transmitting the data to the server, the position of this dynamic observation can be retrieved given the link to its corresponding keyframe. Consequently, the positions of all the dynamic observations benefit from continuous keyframe updates due to \gls{pgo}. To evenly sample paths, we further filter the observations using both hand-crafted heuristics and DBSCAN~\cite{ester1996density} based on timestamps, angles, and relative displacements. This is particularly important for stationary and occluded objects as well as outliers caused by odometry noise. Following an iterative yaw-respective aggregation scheme~\cite{buchner2023learning}, we convert all trajectories into directed graphs, apply Laplacian smoothing, and merge them to build a complete lane graph. Employing the same processing scheme, we add agent trajectories to this graph. Since CURB-SG maintains a connection between the lane graph and the keyframes used in SLAM, we can continuously propagate refinements from \gls{pgo} to the lane graph.
}


{\parskip=3pt
\noindent\textit{Environment Partitioning:} 
Urban outdoor driving scenes exhibit a vastly different topology compared to indoor environments that have been represented using scene graphs so far. We found that classical methods such as wall dilation for retrieving disjoint environment graphs~\cite{hughes2022hydra} are not directly applicable to urban environments. In our work, we propose to separate outdoor environments into intersecting and non-intersecting areas using the obtained lane graph (see above). Ultimately, this gives rise to the hierarchical environment abstraction introduced in CURB-SG enabling efficient querying for downstream tasks such as trajectory prediction.
In particular, we detect intersections based on the following heuristics:
First, we cluster high-degree lane graph nodes to find agglomerations of graph splits and merges. Second, we detect lane graph edges that intersect. These two approaches can be applied to various environments to efficiently handle challenging conditions such as multi-lane roads or non-trivial intersections. After identifying intersection nodes, the remaining disconnected sub-graphs fall into non-intersecting road areas. To assign components from other layers of the scene graph to the extracted partitions, we extend these areas beyond the lane node surroundings as illustrated in \cref{fig:overview}.
}

\section{Experimental Evaluation}

In this section, we evaluate \mbox{CURB-SG} with respect to the collaborative SLAM backend, the constructed lane graph, and the proposed partitioning based on road intersections. 


\subsection{Experimental Setup}

We evaluate CURB-SG on various urban driving scenarios using the CARLA simulator~\cite{dosovitskiy2017carla} due to a lack of real-world multi-agent datasets providing LiDAR scans. In particular, we perform experiments on a set of four diverse environments including \textit{town01}, \textit{town02}, \textit{town07}, and \textit{town10}. Following previous works~\cite{hughes2022hydra}, we use the panoptic annotations with temporally consistent instance IDs provided by the simulator. Where applicable, we demonstrate the efficacy of \mbox{CURB-SG} for one, two, and three agents and average results over ten randomly initialized runs.
Due to the semantics-based voxelization on the server, the total number of map points of a fully explored town is relatively stable. As the path planning of the agents is randomized, it can take a long time until this number is reached. Therefore, we approximate full exploration by using \SI{85}{\percent} as the termination criterion.


\subsection{Collaborative SLAM}

In this section, we evaluate the collaborative SLAM backend of our proposed CURB-SG with respect to both accuracy and cooperative gain in long-term scenarios.


{\parskip=3pt
\noindent\textit{Mapping and Localization}:
In \cref{tab:map-quality}, we present the root mean squared errors (RMSE) of the agents' keyframes and the estimated position of the street signs to represent localization and mapping accuracy, respectively. We compute the position of a street sign as the geometric center of the corresponding bounding box that is inferred from the 3D map. We observe that both errors are reduced when more agents contribute towards the collaborative pose graph. Except for the case of two agents in \textit{town07}, this holds true for the mean as well as the standard deviation across all environments. We further illustrate the robustness of our approach against noisy sensor data by imposing realistic metric Gaussian noise $\mathcal{N}(0,0.02)$ on the LiDAR scans~\cite{velodyne_noise} of the agents in \textit{town01} and \textit{town02}. As shown in \cref{tab:map-quality}, the noise does not significantly alter the errors indicating that downstream tasks such as lane graph estimation do not degrade either.} 

\begin{table}
\scriptsize
\centering
\caption{Evaluation of localization and mapping performance}
\vspace{-0.2cm}
\label{tab:map-quality}
\setlength\tabcolsep{3.0pt}
\begin{threeparttable}
    \begin{tabular}{cc | c c c}
        \toprule
        \multirow{2}{*}{Environment} & Agent & RMSE (agents) & RMSE (street signs) & Exploration time\\
         & count & [m] & [m] & [sim. steps] \\
        \midrule
        \multirow{3}{*}{\textit{town01}} & 1 & 0.735 $\pm$ 0.492 & 0.865 $\pm$ 0.485 & 2502.50\\
        & 2 & 0.368 $\pm$ 0.343 & 0.480 $\pm$ 0.358 & 1267.70\\
        & 3 & \textbf{0.132 $\pm$ 0.096} & \textbf{0.169 $\pm$ 0.096} & \textbf{1134.40}\\
        \greyrule
        \textit{ + noise} & 3 & 0.159 $\pm$ 0.093 & 0.225 $\pm$ 0.101 & --\\
        \midrule
        \multirow{3}{*}{\textit{town02}}& 1 & 0.306 $\pm$ 0.208 & 0.297 $\pm$ 0.194 & 2079.00\\
        & 2 & 0.249 $\pm$ 0.176 & 0.299 $\pm$ 0.238 & 943.80\\
        & 3 & \textbf{0.126 $\pm$ 0.109} & \textbf{0.164 $\pm$ 0.148} & \textbf{597.60}\\
        \greyrule
        \textit{ + noise} & 3 & 0.119 $\pm$ 0.078 & 0.140 $\pm$ 0.064 & --\\
        \midrule
        & 1 & 0.564 $\pm$ 0.406 & -- & 3632.70\\
        \textit{town07}
        & 2 & 0.234 $\pm$ \textbf{0.182} & -- & 1760.20 \\
        & 3 &\textbf{0.218} $\pm$ 0.198 & -- & \textbf{910.00}\\
        \midrule
        & 1 & 0.333 $\pm$ 0.185 & -- & 923.70\\
        \textit{town10}
        & 2 & 0.310 $\pm$ 0.185 & -- & 724.00\\
        & 3 & \textbf{0.116 $\pm$ 0.106} & -- & \textbf{391.10} \\
        \bottomrule
    \end{tabular}
    \footnotesize
    Mean and standard deviation over ten runs of the RMSE of the agents' keyframes and the estimated position of the street signs representing localization and mapping accuracy, respectively. Note that the environments \textit{town07} and \textit{town10} do not contain street signs. The rightmost column lists the mean time required to map \SI{85}{\percent} of the entire town measured in simulation steps. 
\end{threeparttable}
\end{table}

\begin{figure}
    \centering
    \includegraphics[width=0.95\linewidth]{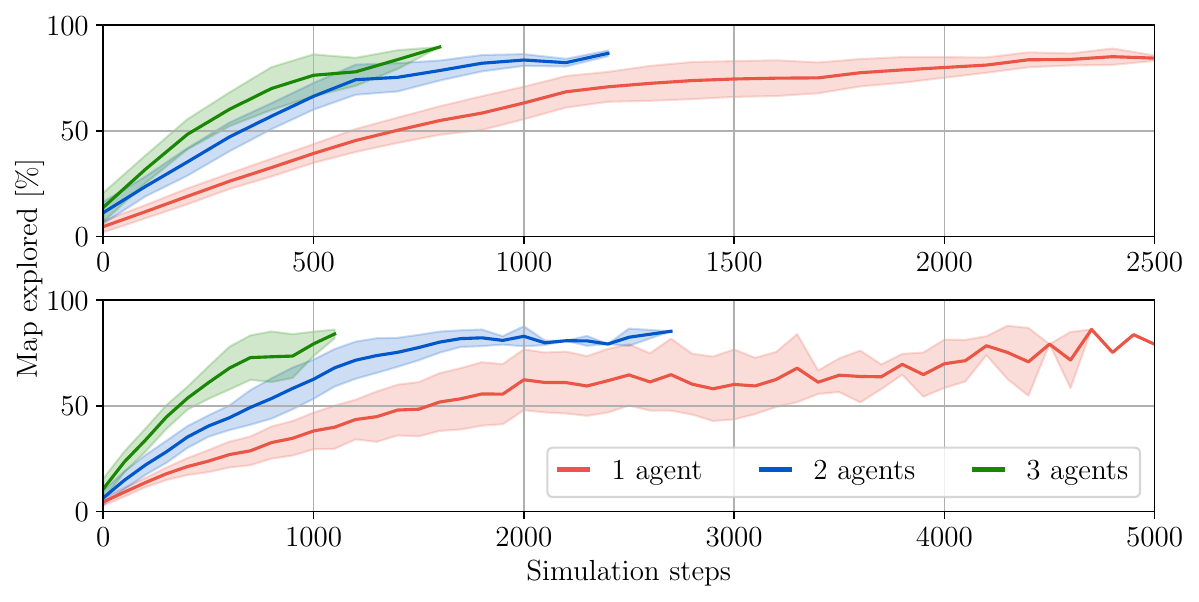}
    \vspace{-.3cm}
    \caption{The mapping progress in \textit{town02} (top) and \textit{town07} (bottom) for one, two, and three agents. Our collaborative SLAM method benefits from receiving inputs from multiple agents.
    }
    \label{fig:exploration-time}
    \vspace{-0.3cm}
\end{figure}
\begin{figure}
    \centering
    \includegraphics[width=.95\linewidth]{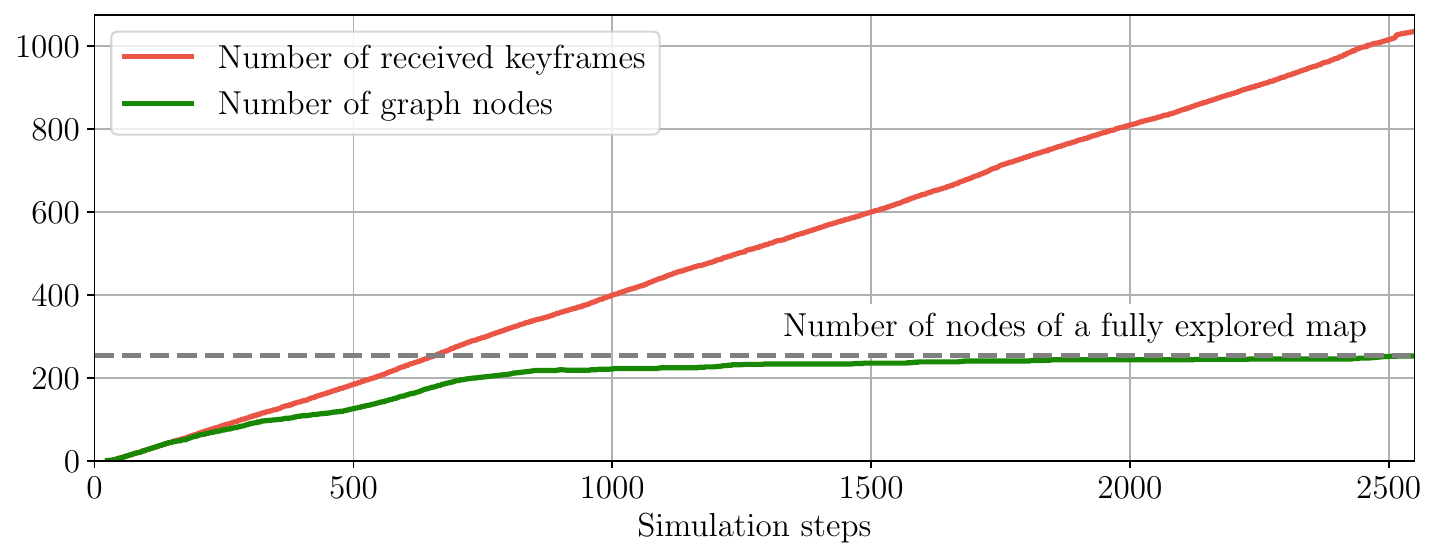}
    \vspace{-.1cm}
    \caption{Our proposed edge contraction mechanism effectively reduces the number of nodes in the pose graph to maintain the capability of frequent graph optimization. This plot shows three agents operating in \textit{town02}.}
    \label{fig:nodes-over-time}
    \vspace{-0.1cm}
\end{figure}

{\parskip=3pt
\noindent\textit{Long-Term Mapping:}
We demonstrate the efficacy of our proposed adaptions of HDL Graph SLAM~\cite{koide2019a} (see \cref{ssec:collaborative-slam}) to address long-term mapping of large areas.
In the rightmost column of \cref{tab:map-quality}, we report the time required to map a town when using one, two, or three agents. Generally, the higher the number of contributing agents, the smaller the time required to explore the map.
Similarly, in \cref{fig:exploration-time}, we illustrate the mapping progress measured by the number of 3D points versus the simulation steps. While the results confirm the aforementioned general trend towards faster exploration in a multi-agent setup, the pure mapping speed will reach an upper bound above that additional agents will not further increase the speed. However, even afterward, these agents will keep sending measurements and vehicle observations contributing towards frequent map updates and enhancing the lane graph (see \cref{ssec:exp-lane-graph}). We present further results for \textit{town01} and \textit{town10} in the suppl. material Sec.~S.3.\looseness=-1

Finally, we demonstrate that our proposed edge contraction successfully limits the number of nodes contained in the pose graph. In \cref{fig:nodes-over-time}, we show the example of three agents operating in \textit{town02} and compare the number of optimizable graph nodes with the total number of keyframes sent by the agents. We observe that without edge contraction, the pose graph continuously grows with the number of keyframes sent, rendering frequent optimization infeasible.
}

\subsection{Lane Graph}
\label{ssec:exp-lane-graph}

\begin{figure}[t]
    \centering
    \includegraphics[width=.45\linewidth]{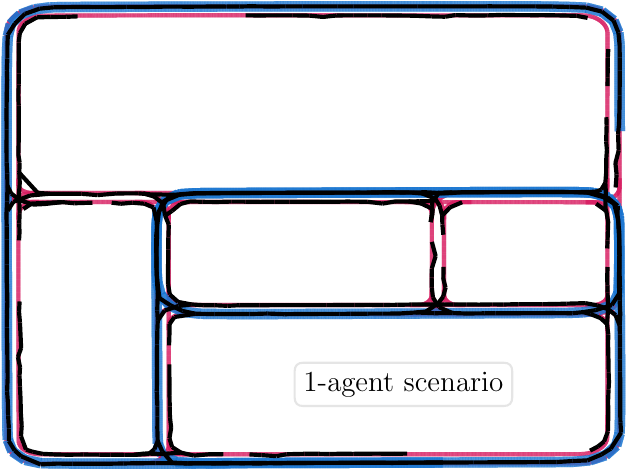}
    \hfill
    \includegraphics[width=.45\linewidth]{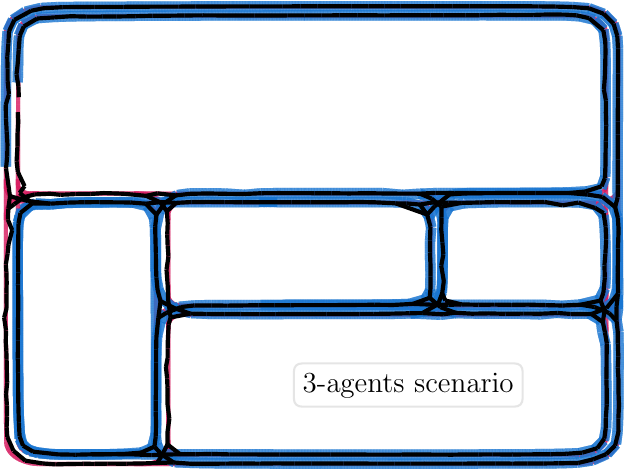}
    \includegraphics[width=.80\linewidth]{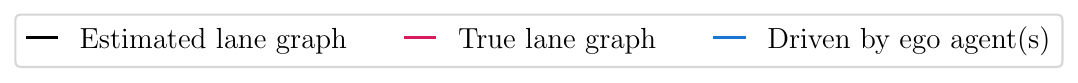}
    \vspace{-.3cm}
    \caption{Visualization of the constructed lane graph of \textit{town02} when using one or three agents. Lanes marked in blue have been traversed by an ego agent. Others are reconstructed from observing surrounding vehicles.}
    \label{fig:lane-graph}
    \vspace{-0.3cm}
\end{figure}

We evaluate our proposed online lane graph generation approach from the paths of the ego agents and their observations of other vehicles (\cref{ssec:scene-graph}). We present qualitative results in \cref{fig:lane-graph} for two scenarios simulated in \textit{town02} with 30 additional non-agent vehicles: the left figure visualizes the lane graph in a single-agent scenario terminated as soon as the agent starts to repeatedly revisit intersections. Although the path of the agent, shown in blue, does not cover all the lanes, including the paths of the observed vehicles allows for a substantial extension of the lane graph. The right figure depicts a long-term scenario with three agents demonstrating that collaboration further boosts performance. Our method yields an almost complete lane graph even though several lanes have only been driven by the agents in the opposite direction.\looseness=-1

We quantify these findings in \cref{tab:lane-graph} following previous works on lane graphs: precision and recall of the TOPO and GEO metrics~\cite{he_wacv}, APLS~\cite{etten2018spacenet}, SDA\textsubscript{R}~\cite{buchner2023learning} with the subscript denoting the search radius in meters, and the graph IoU~\cite{buchner2023learning}. We observe that except for the TOPO/GEO precision and the APLS in the 3-agent scenario, all the metrics show an improvement when using not only the paths of the ego agents but also of the observed vehicles. We attribute the decrease in precision to the noise in the estimated position of the other vehicles. Since we approximate the center of a vehicle by the geometric mean of the respective 3D points, there is a bias towards the center line of a road for all oncoming cars. We further observe that increasing the number of agents does have a positive impact on all the metrics except for the TOPO/GEO precision and the SDA\textsubscript{4.5} demonstrating the efficacy of our method.

\begin{table}
\scriptsize
\centering
\caption{Lane graph evaluation}
\vspace{-0.2cm}
\label{tab:lane-graph}
\setlength\tabcolsep{3.3pt}
\begin{threeparttable}
    \begin{tabular}{cc | cccccc }
        \toprule
        Ego & Obs. & TOPO P / R & GEO P / R & APLS & SDA\textsubscript{4.5} & SDA\textsubscript{9.0} & Graph IoU \\
        \midrule
        \multicolumn{3}{l}{\textbf{1-agent scenario}} \\
        [.5ex]
        \cmark & & \textbf{0.810} / 0.281 & \textbf{0.923} / 0.415 & 0.658 & 0.000 & 0.042 & 0.386 \\
        \cmark & \cmark & 0.678 / \textbf{0.562} & 0.855 / \textbf{0.812} & \textbf{0.724} & \textbf{0.278} & \textbf{0.394} & \textbf{0.690} \\
        \midrule
        \multicolumn{3}{l}{\textbf{3-agents scenario}} \\
        [.5ex]
        \cmark & & \textbf{0.715} / 0.583 & \textbf{0.874} / 0.762 & \textbf{0.800} & 0.188 & 0.357 & 0.658 \\
        \cmark & \cmark & 0.574 / \textbf{0.712} & 0.751 / \textbf{0.925} & 0.756 & \textbf{0.250} & \textbf{0.452} & \textbf{0.751} \\
        \bottomrule
    \end{tabular}
    \footnotesize
    Quantitative results obtained in \textit{town02}. The two left columns indicate whether only the paths of the ego agents or also the estimated positions of other observed vehicles have been used. For the TOPO and GEO metrics, we provide both precision (P) and recall (R).
\end{threeparttable}
\vspace{-0.1cm}
\end{table}

\subsection{Environment Partitioning}
\label{ssec:env-part}
\begin{figure}
    \centering
    \includegraphics[width=0.95\linewidth]{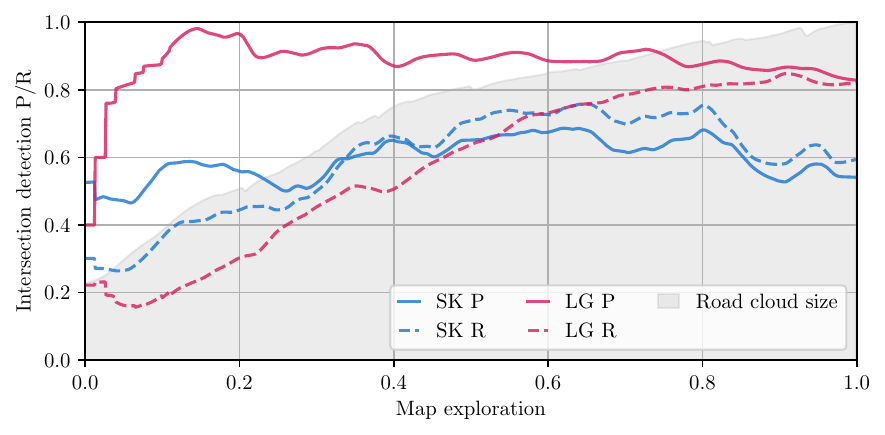}
    \vspace{-.4cm}
    \caption{Intersection detection quality of our lane graph-based detection of intersections (LG) and an image-based skeletonization baseline of the road surface (SK). Average precision (P) and recall (R) of both approaches across 10 runs with 3 agents and 40 vehicles on \textit{town02} as well as the size of the investigated road surface point cloud are shown.}
    \label{fig:intersections-over-time}
    \vspace{-0.4cm}
\end{figure}

We evaluate our approach for environment partitioning (\cref{ssec:scene-graph}) by comparing it against the ground-truth intersection points of the underlying map. Throughout exploring the environment, the recall is normalized using the point cloud of the road surface obtained thus far. Our proposed lane graph-based method (LG) is compared against a morphological image skeletonization baseline (SK) that uses medial axes of the bird's-eye-view projected point cloud of the road surface. Kernelized smoothing and dilation followed by thresholding the obtained bird's-eye-view image helps in filtering false positive points and noise. In order to further increase precision, the SK baseline includes clustering culmination of intersection points in local areas that originate from artifacts in the skeleton graph.
We report the precision and recall values across ten exploration runs on \textit{town02} in \cref{fig:intersections-over-time}. We observe that our approach (LG) achieves at least \SI{20}{\percent} greater precision while showing comparable or exceeding recall scores. As our approach relies on observed vehicle trajectories, we attribute the lower initial recall of the LG method to a small number of initially seen trajectories while the point cloud-based baseline already processes a larger extent of the surroundings at this stage. Nonetheless, we observe that the SK baseline yields vastly different partitioning solutions throughout exploration as it is not robust to artifacts such as occlusions due to vehicles or sparse LiDAR readings of distant road surfaces. We believe that a conservative, high-precision classifier is beneficial as over-segmentation increases the number of roads and intersections unnecessarily. Further explanations are provided in suppl. material Sec.~S.4. Additionally, we observe that simply extracting intersections from the pose graph produces low recalls as every path has to be traversed by the agents instead of relying on more descriptive observations.

\section{Conclusion}

In this work, we introduced CURB-SG as a novel approach to building large-scale hierarchical dynamic 3D urban scene graphs from multi-agent observations. We furthermore demonstrated how our collaborative SLAM approach facilitates frequent map updates and rapid exploration while scaling to large environments. To foster further research in this direction, we made our code publicly available. In future work, we will address the reliance on simulated panoptic labels and known initial poses of the agents.
Orthogonal to that, follow-up work could address a decentralized variant that operates under real-time constraints. Furthermore, we plan to include pedestrian information as well as additional topological elements such as road boundaries.


\begin{footnotesize}
    \bibliographystyle{IEEEtran}
    \bibliography{references.bib}
\end{footnotesize}


\clearpage
\renewcommand{\baselinestretch}{1}
\setlength{\belowcaptionskip}{0pt}

\begin{strip}
\begin{center}
\vspace{-5ex}

\textbf{\LARGE \bf
\includegraphics[width=\textwidth]{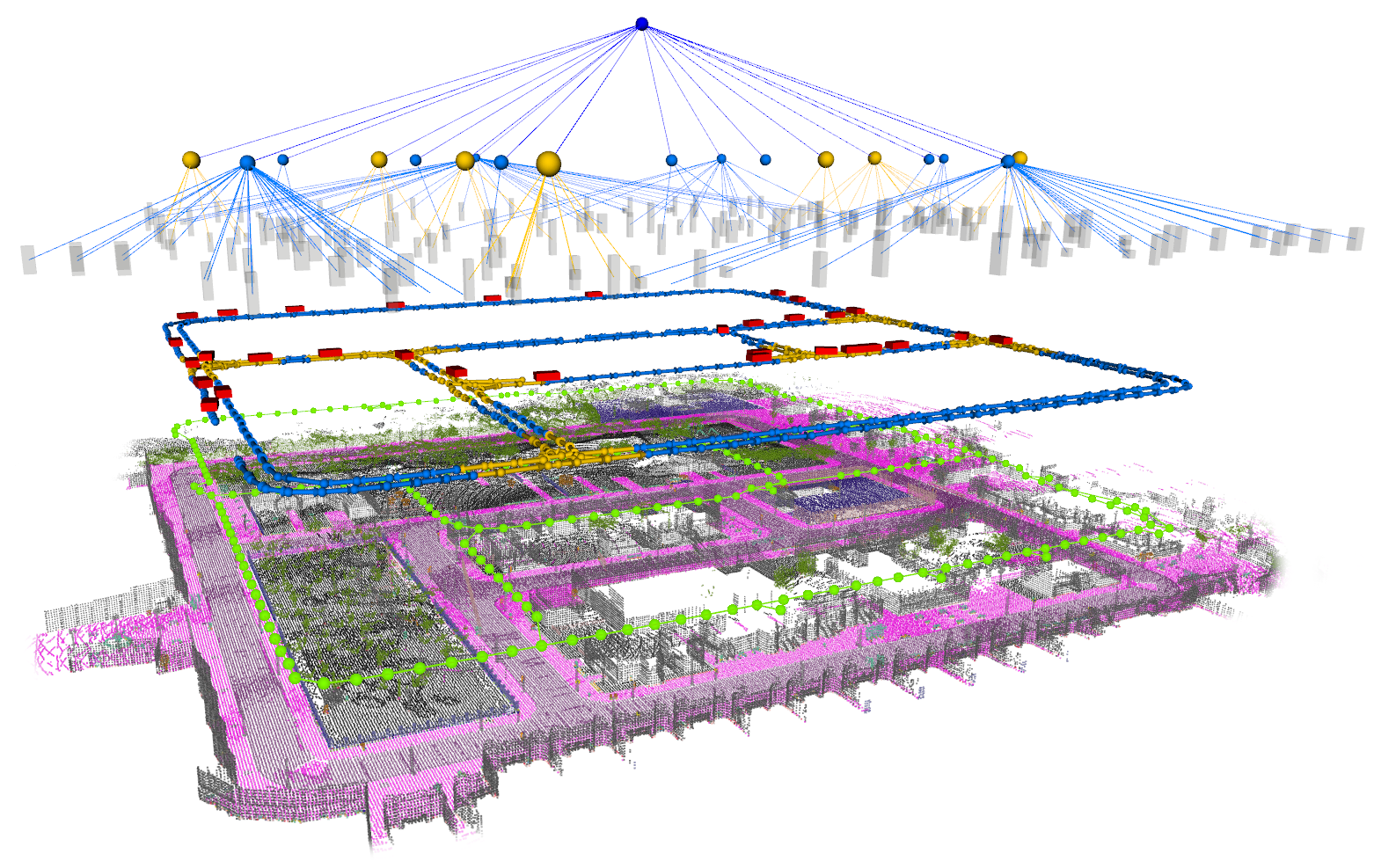} \\[1cm]
Collaborative Dynamic 3D Scene Graphs for Automated Driving} \\
\vspace{3ex}

\Large{\bf- Supplementary Material -}\\
 \vspace{0.4cm}
 \normalsize{Elias Greve$^{1*}$,
Martin Büchner$^{1*}$,
Niclas Vödisch$^{1*}$,
Wolfram Burgard$^{2}$,
and Abhinav Valada$^{1}$}
\end{center}
\end{strip}

\setcounter{section}{0}
\setcounter{equation}{0}
\setcounter{figure}{0}
\setcounter{table}{0}
\setcounter{page}{1}
\makeatletter

\renewcommand{\thesection}{S.\arabic{section}}
\renewcommand{\thesubsection}{S.\arabic{subsection}}
\renewcommand{\thetable}{S.\arabic{table}}
\renewcommand{\thefigure}{S.\arabic{figure}}


 \let\thefootnote\relax\footnote{$^{*}$ Equal contribution.\\
 $^{1}$ Department of Computer Science, University of Freiburg, Germany.\\
 $^{2}$ Department of Eng., University of Technology Nuremberg, Germany.
 }%
\normalsize


In this supplementary material, we provide additional information on the implementation details of our proposed approach CURB-SG as well as a number of insights regarding our experimental evaluation. This ranges from our semantic voxel grid filtering and collaborative SLAM to our lane graph-based environment partitioning method.
For convenience, we follow the structure of the main manuscript.


\section{Implementation Details}

\subsection{Simulation Environment}

\begin{figure}[ht]
    \centering
    \captionsetup[subfigure]{justification=centering}
    \subfloat[\textit{town01}]{%
        \includegraphics[width=.485\linewidth]{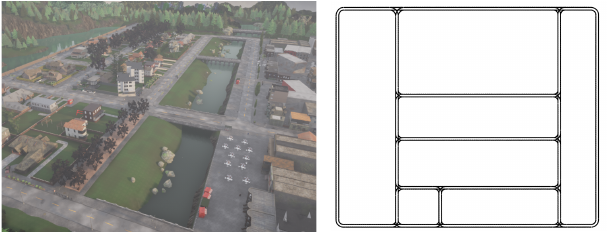}}
    \hfill
    \subfloat[\textit{town02}]{%
        \includegraphics[width=.485\linewidth]{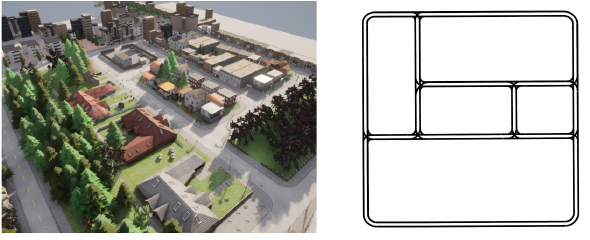}}
    \vfill
    \vspace{-.2cm}
    \subfloat[\textit{town07}]{%
        \includegraphics[width=.485\linewidth]{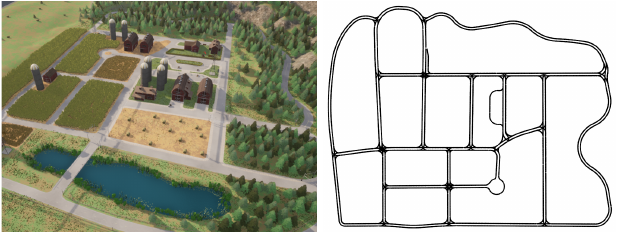}}
    \hfill
    \subfloat[\textit{town10}]{%
        \includegraphics[width=.485\linewidth]{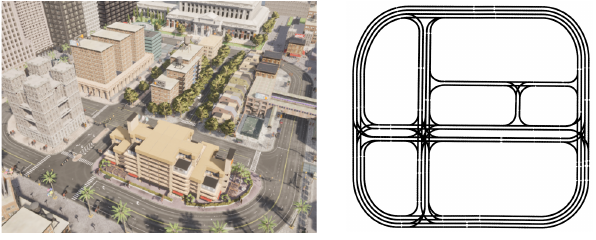}}
    \caption{We use four urban environments from the CARLA simulator~\cite{dosovitskiy2017carla}.}
    \label{fig-suppl:carla-towns}
\end{figure}

Throughout our evaluation, we make use of four different CARLA map environments as shown in \cref{fig-suppl:carla-towns} that contain classical two-way roads (\textit{town01, town02, town07}), roundabouts (\textit{town07}) as well as bidirectional multi-lane roads (\textit{town10}). We spawn 30 to 40 additional non-agent vehicles to simulate traffic behavior. The navigation of all agents and passive vehicles is random and for the agent-only evaluations, e.g., localization and map exploration, all the traffic lights are constantly green to avoid unnecessary waiting time for the agents.

We simulate a semantic $360^{\circ}$ LiDAR sensor using the CARLA Python API with the following parameters: 32 channels, \SI{2.8}{\meter} above ground surface, \SI{10}{\meter} minimum distance, \SI{10}{\hertz} rotation frequency, 56,000 points per second, azimuth range of [$-30^{\circ}, 10^{\circ}$]. Our setup runs ROS Noetic and CARLA 0.9.13 on an Intel i5-6500 @ 4 x \SI{3.2}{\giga\hertz} and an NVIDIA TITAN X GPU. The implementation uses both C++~(g++ 9.4.0) and Python~(3.8).

\section{Semantic Voxel Grid Filtering}
\begin{figure}
    \centering
    \includegraphics[width=\linewidth]{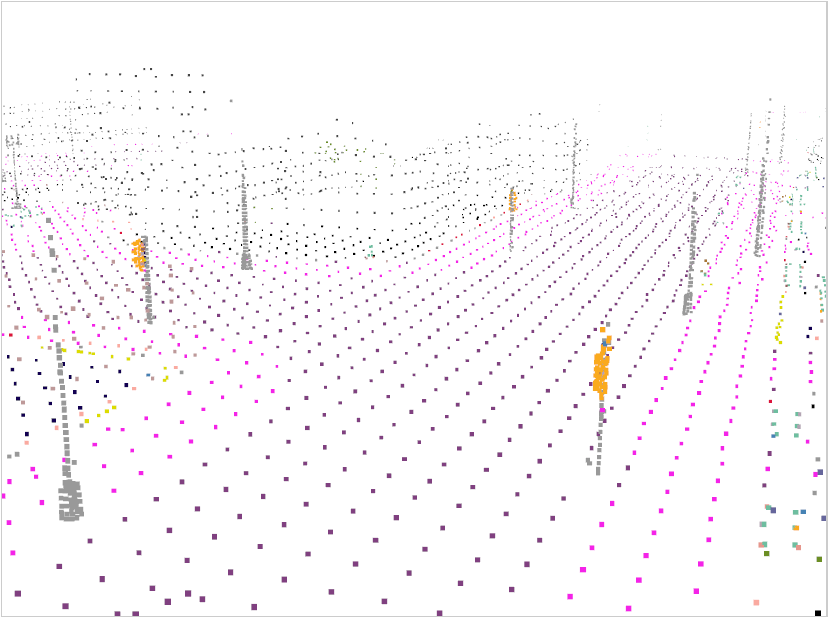}
    \vspace{-.4cm}
    \caption{Point cloud visualization of the obtained panoptic point cloud after semantics-based voxel grid filtering. As shown, typically small classes such as traffic signs and poles show a finer resolution while, e.g., the \textit{road} class is filtered with a coarser grid.}
    \label{fig-suppl:semantic_voxelgrid}
\end{figure}

As briefly described in \cref{ssec:collaborative-slam}, we adapt the original implementation of HDL Graph SLAM~\cite{koide2019a} regarding the prefiltering of point clouds. Instead of using a class-agnostic voxel grid filter, our semantic class-respective filter uses different grid sizes for different semantic classes. Therefore, smaller objects such as landmarks are represented using a higher number of points while typical ``stuff'' classes such as \textit{road} use fewer points, as shown in \cref{fig-suppl:semantic_voxelgrid}. The class-respective voxel grid resolutions are given in \cref{tab-suppl:voxel_grid_res}.

\begin{table}[t]
\scriptsize
\centering
\caption{Semantic voxel grid resolution.}
\vspace{-0.2cm}
\label{tab-suppl:voxel_grid_res}
\begin{threeparttable}
    \begin{tabular}{l | l}
        \toprule
        Resolution [m] & Semantic class\\
        \midrule
        0.15 (fine) & \textit{Fence}, \textit{Pole}, \textit{TrafficSign}, \textit{GuardRail}, \textit{TrafficLight}, \textit{Static}, \textit{Dyn.} \\
        \midrule
        \multirow{3}{*}{0.5 (coarse)} & \textit{Unlabeled}, \textit{Building}, \textit{Other}, \textit{Pedestrian}, \textit{Road}, \textit{SideWalk}, \\
        & \textit{Vegetation}, \textit{Vehicles}, \textit{Wall}, \textit{Sky}, \textit{Ground}, \textit{Bridge}, \textit{RailTrack}, \\
        & \textit{Water}, \textit{Terrain}  \\
        \bottomrule
    \end{tabular}
    \footnotesize
    All semantic classes given by CARLA and the respective voxel grid resolution utilized throughout our experiments.
\end{threeparttable}
\end{table}

\begin{figure}
    \centering
    \captionsetup[subfigure]{justification=centering}
    \subfloat[RMSE of the agent localization of two runs with different filter approaches.]{%
        \includegraphics[width=\linewidth]{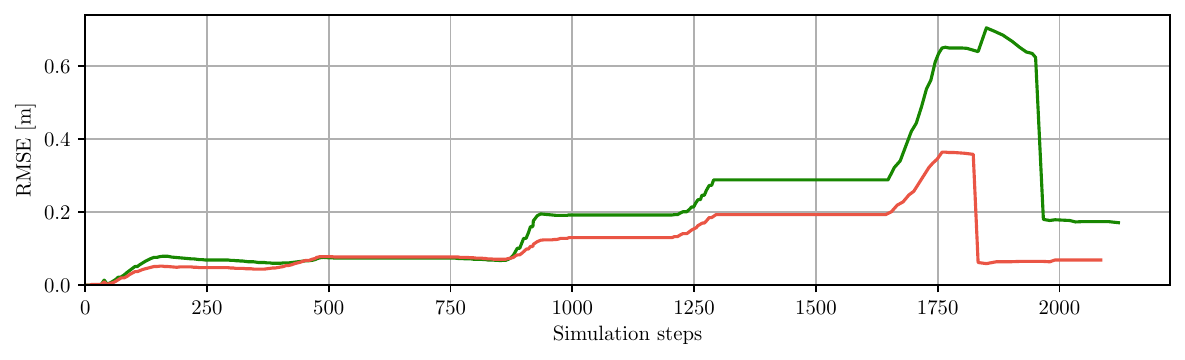}}
    \vfill
    \vspace{-.2cm}
    \subfloat[RMSE of the traffic signs of two runs with different filter approaches.]{%
        \includegraphics[width=\linewidth]{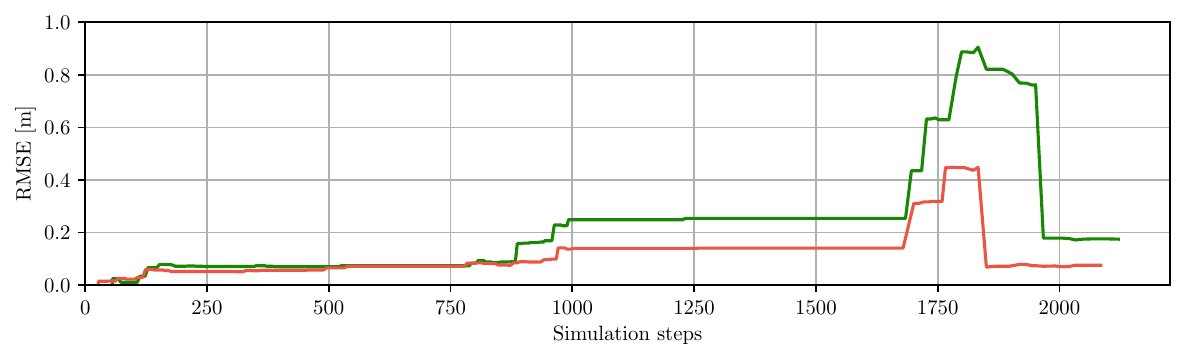}}
    \vfill
    \vspace{.2cm}
    \includegraphics[width=0.7\linewidth]{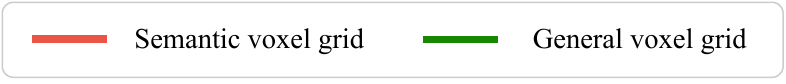}
    \caption{Comparison of a general voxel grid filter to the semantic voxel grid filter.}
    \label{fig-suppl:supp_filter_comp}
\end{figure}

This approach not only bears the potential to decrease the size of the stored point clouds but also leads to increased localization accuracy. As depicted in \cref{fig-suppl:supp_filter_comp}, we observe smaller localization errors regarding both the agents as well as, e.g., traffic signs compared to the generic voxel grid filter. The shown data was collected by letting the agent go counter-clockwise along the outside road of \textit{town02} with a \SI{15}{\meter} segment overlap at the end. Since a loop closure is detected at the end of the run, we observe a sudden decrease in the root mean squared error. After 2000 simulation steps, this induces a $\sim 50\%$ improvement using the semantic compared to the standard voxel grid filter.

\section{Collaborative SLAM}
Complementary to the findings in \cref{fig:exploration-time}, we provide the exploration performance for one, two, and three agents on \textit{town01} and \textit{town10}. Similarly, we observe that three agents show the smallest time needed for 85\% map exploration. However, as shown for \textit{town01}, the effect of a third agent may potentially vanish on certain maps. Nonetheless, in an open-world setting, we believe that more agents lead to faster and more accurate mapping results as already shown in \cref{tab:map-quality}.

\section{3D Urban Scene Graphs}

\begin{figure*}
    \centering
    \includegraphics[width=\textwidth]{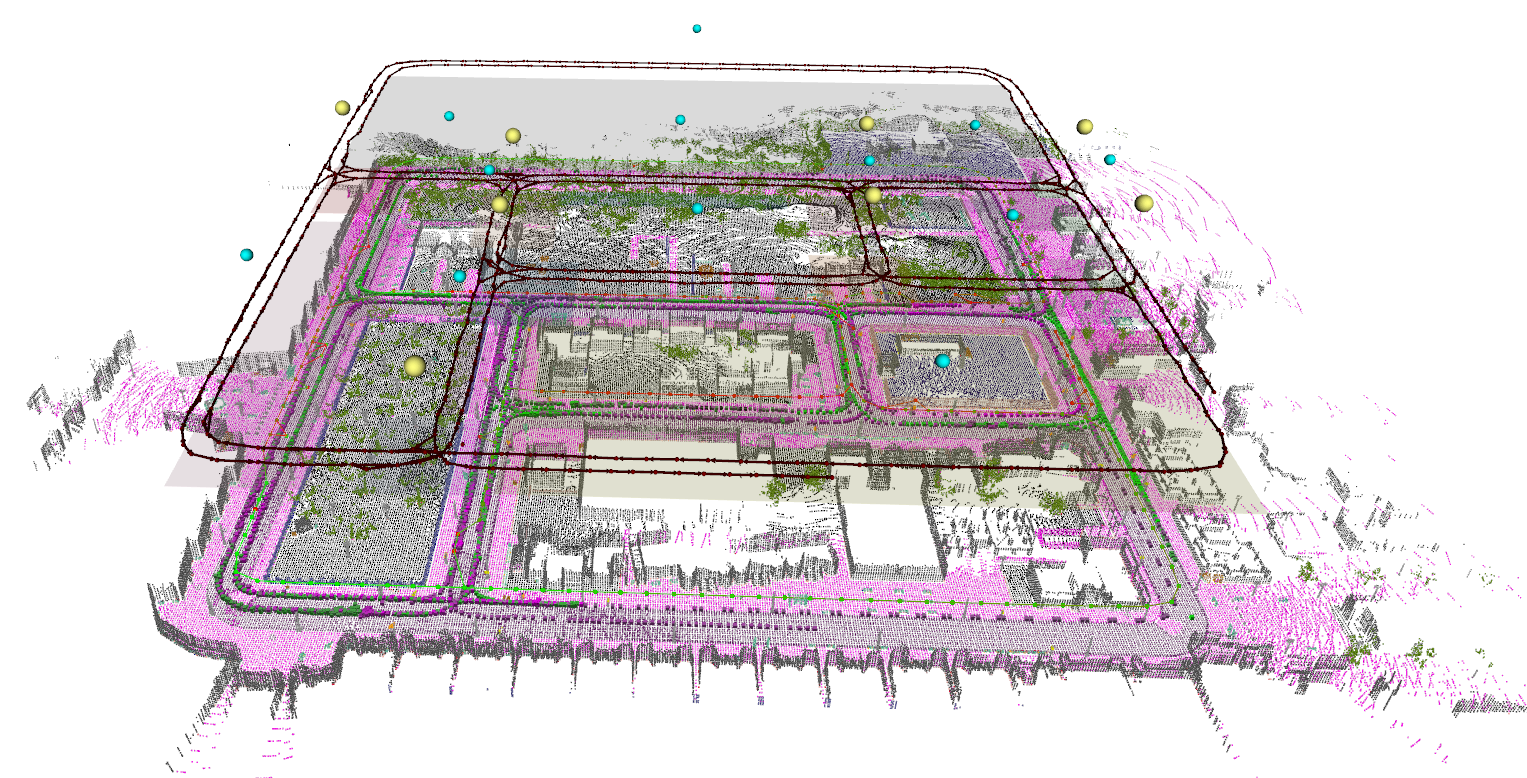}
    \vspace{-.5cm}
    \caption{Qualitative visualization of the intersection detection quality. Yellow nodes represent intersections, whereas blue nodes represent non-intersecting road areas. As shown, these areas can be predicted with high precision even without a complete lane graph.}
    \label{fig-suppl:intersection_detection}
\end{figure*}

\subsection{Environment Partitioning}
\begin{figure}
    \centering
    \includegraphics[width=\linewidth]{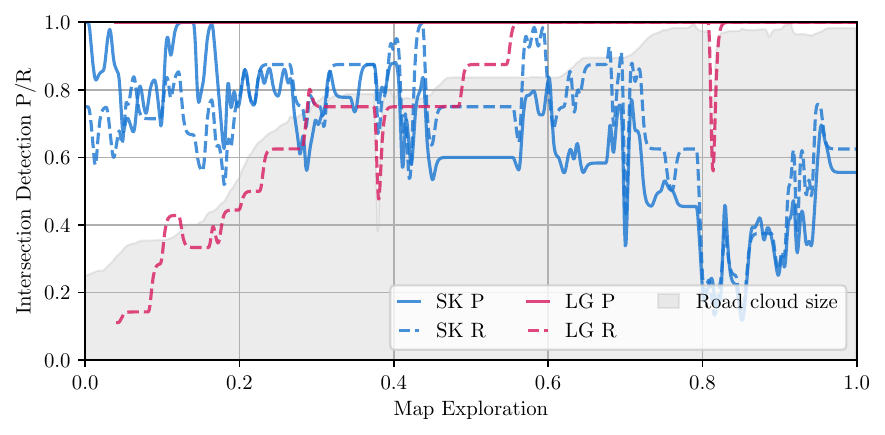}
    \caption{Intersection detection quality of a single run on map \textit{town02}. Precision (P, solid) and recall (R, dashed) of both our lane graph-based method (LG) and the skeletonization baseline (SK) are plotted against continuous exploration of the map.}
    \label{fig-suppl:intersection_detection_single}
\end{figure}

In addition to our findings in \cref{ssec:env-part}, we provide further qualitative and quantitative findings. Different from \cref{fig:intersections-over-time}, in \cref{fig-suppl:intersection_detection_single}, we show the intersection detection precision (P) and recall (R) for a single exploration run instead of the average across 10 runs. While the precision of the lane graph-based method (LG) stays constantly at 1.0 the recall generally increases throughout exploration. Minor decreases in the recall score are compensated immediately. In contrast, the image-based skeletonization method (SK) shows highly volatile precision and recall curves that do not improve while greater extents of the environment are explored. In general, this leads to a vast number of different segmentation solutions while exploring, which is generally not suited for robust mapping.

In \cref{fig-suppl:intersection_detection}, we portray the qualitative output of CURB-SG on \textit{town02}. A directed graph colored in black represents the lane graph, whereas a green graph slightly above the semantic point cloud denotes the obtained pose graph that is subject to (inter-agent) loop closures.
Yellow-colored nodes above the lane graph denote intersections while blue nodes represent non-intersecting road segments. Although the produced lane graph shows slight inaccuracies and is not fully complete, we can segment areas under both high precision and recall.

{\parskip=3pt
\noindent\textit{Morphological Intersection Detection Baseline}:
In the following, we provide additional insights on the morphological partitioning baseline with which we compare in the main manuscript. To singulate intersection points we take the obtained road surface point cloud and project it to the bird's-eye-view space, which is represented as an image using the min/max coordinates of the road points. In the next step, we apply $(5,5)$-kernelized smoothing in the 2D domain. Additionally, we dilate the obtained surface using a $(9,9)$-sized kernel followed by thresholding to fill holes created by dynamic object occlusions. To obtain the road connectivity, we extract the medial axes of the obtained surface using \texttt{scikit-image} and turn it into a sparse graph that is used for evaluation.
}

{\parskip=3pt
\noindent\textit{Intersection Detection Accuracy}:
We obtain the intersection recall, precision, and accuracy by calculating the pair-wise Euclidean distances among all ground-truth intersection positions. We then solve the linear sum assignment problem to associate predictions with ground-truth intersection points. Finally, we compute the intersection metrics based on the obtained assignment.
}

\begin{figure}
    \centering
    \includegraphics[width=\linewidth]{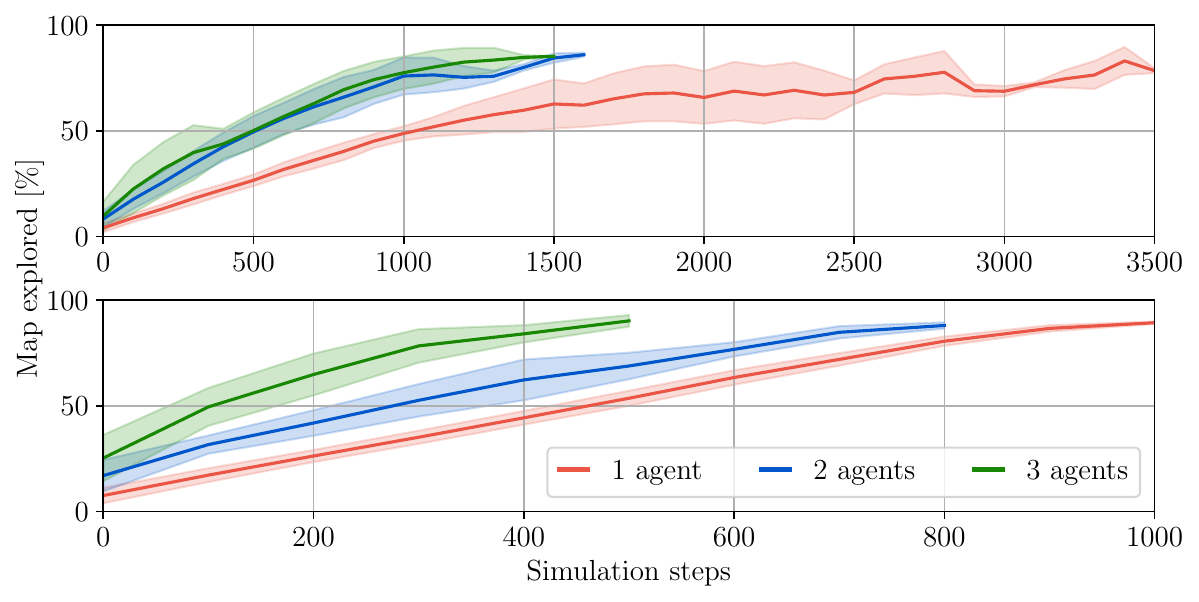}
    \caption{The mapping progress in \textit{town01} (top) and \textit{town10} (bottom) for one, two, and three agents. Our collaborative SLAM method benefits from receiving inputs from multiple agents. Although additional agents might not further increase the mapping speed, they will continue contributing towards frequent map updates and refining the lane graph.}
    \label{fig-suppl:exploration-time}
\end{figure}

\end{document}